# Subasa - Adapting Language Models for Low-resourced Offensive Language Detection in Sinhala

**Shanilka Haturusinghe**♣, **Tharindu Cyril Weerasooriya**♢, **Marcos Zampieri**♠,

**Christopher M. Homan**♢, **S.R. Liyanage**♣[*]

♣University of Kelaniya, Sri Lanka, ♢Rochester Institute of Technology, USA

♠George Mason University, USA

s.haturusinghe99@gmail.com, {tw3318,mazgla,cmhvcs}@rit.edu, sidath@kln.ac.lk

## Abstract

This paper contains expressions that may offend the readers.

Accurate detection of offensive language is essential for a number of applications related to social media safety. There is a sharp contrast in performance in this task between low- and high-resource languages. In this paper, we adapt fine-tuning strategies that have not been previously explored for Sinhala in the downstream task of offensive language detection. Using this approach, we introduce four models: "Subasa-XLM-R", which incorporates an intermediate Pre-Finetuning step using Masked Rationale Prediction. Two variants of "Subasa-Llama" and "Subasa-Mistral", are fine-tuned versions of Llama (3.2) and Mistral (v0.3), respectively, with a task-specific strategy. We evaluate our models on the SOLD benchmark dataset for Sinhala offensive language detection. All our models outperform existing baselines. Subasa-XLM-R achieves the highest Macro F1 score (0.84) surpassing state-of-the-art large language models like GPT-4o when evaluated on the same SOLD benchmark dataset under zero-shot settings. The models and code are publicly available.[1]

## 1 Introduction

A major challenge in the field of NLP are the disparities between high- and low-resource languages. These impact foundational language models as well as downstream tasks such as offensive language detection (Weerasooriya et al., 2023a), an important task at the intersection of social media analysis and NLP.

As people increasingly spend a significant portion of their day on online platforms like social media, their exposure to offensive or abusive language has surged (Bertaglia et al., 2021). This trend is equally visible in Sri Lanka, where a substantial amount of social media content is generated in Sinhala. Studies show that an alarming amount of this content is hateful, and the severity of this issue is evident from several instances in recent years where the Sri Lankan government had to block social media platforms entirely to curb its spread, as it had fueled real-world unrest (Awais et al., 2020).

Sinhala (සිංහල) is an Indo-Aryan language spoken by over 17 million people in Sri Lanka and remains a low-resource language (De Silva, 2019). For offensive language detection specifically, systems for Sinhala lag behind those developed for resource-rich languages like English, Spanish, and Mandarin (Avetisyan and Broneske, 2023; Ranasinghe et al., 2024). To the best of our knowledge, fewer than five annotated offensive language datasets exist for Sinhala, demonstrating its status as a low-resource language (Ranasinghe et al., 2024).

While state-of-the-art large language models (LLM) like GPT-4o demonstrate strong performance in many languages, our evaluations suggest they struggle to reliably identify offensive language in Sinhala (results detailed in Section 4). At the time of submission, the Perspective API (Lees et al., 2022) which is utilized extensively in both academia and industry for the purpose of identifying offensive content does not provide support for Sinhala.

Our work addresses these shortcomings by introducing *Subasa* ("සුබස"), which translates to *wholesome language*. In this paper, we present four variants of Subasa. These models improve the current state of offensive language detection for Sinhala by adapting fine-tuning strategies previously unexplored for Sinhala.

We address the following research questions:

[*]Shanilka Haturusinghe is the primary author. S.R. Liyanage is the Corresponding Author.

[1]Access code and models via https://github.com/haturusinghe/subasa-llm and https://github.com/haturusinghe/subasa-plm

**RQ1**: Can intermediate pre-finetuning tasks—specifically masked rationale prediction (MRP)—effectively improve pre-trained language models (PLMs) for offensive language detection in Sinhala?

**RQ2**: Can task-specific fine-tuning strategies improve the effectiveness of LLMs for offensive language detection in Sinhala?

## 2 Related Work

Shared tasks like TRAC (Kumar et al., 2018) and HASOC (Chakravarthi et al., 2021) have established offensive language detection as an important NLP challenge, yet progress remains unevenly distributed across languages. Generally, building an effective model for offensive language detection is challenging due to the subjective nature of what constitutes offensive content, which can vary based according to individual beliefs (Weerasooriya et al., 2023b). Most research has focused on high-resource languages like English, French, German, and Spanish, benefiting from the availability of large datasets (Zampieri et al., 2022). In contrast, research on low-resource languages highlights the difficulties in detecting offensive language (Mozafari et al., 2022), with notable studies in Tamil (Balakrishnan et al., 2023), Arabic (Shannag et al., 2022), South African languages (Oriola and Kotzé, 2020) and also for Sinhala (Dias et al., 2018; Fernando et al., 2022; Munasinghe and Thayasivam, 2022).

Pretrained language models (PLM) have emerged as a powerful approach for a number of NLP tasks including offensive language detection. BERT variants have shown success when fine-tuned for this task across both high-resource languages like English (Jahan et al., 2021) and lower-resource contexts like Arabic (Althobaiti, 2022) and Sinhala (Rajapaksha et al., 2023). While intermediate task training has shown promise in enhancing PLM performance across various NLP tasks—from semantic parsing (Pruksachatkun et al., 2020) to natural language understanding (Aghajanyan et al., 2021)—its application to offensive language detection emerged only recently with the introduction of Masked Rationale Prediction (MRP) by Kim et al. (2022). Though MRP demonstrated significant improvements for English, its potential remains unexplored for low-resource languages. We are the first to adapt MRP to Sinhala, addressing the language's data scarcity.

LLMs are transformer-based models with billions of parameters trained on massive training corpora (Chowdhery et al., 2023). While LLMs perform well in high-resource languages like English, their effectiveness in low-resource languages is often limited, as highlighted in various studies (Ahuja et al., 2023). Adapting LLMs for low-resource languages is challenging because most are pre-trained primarily on English data. Approaches to address this include; (i) continuing training with non-English data, (ii) transferring knowledge via supervised fine-tuning, and (iii) extending the LLMs vocabulary to include non-English tokens (Toraman, 2024). For instance, Toraman (2024) demonstrated that fine-tuned LLMs can achieve strong performance even with limited data, as shown for Turkish. Jayakody and Dias (2024) evaluated the GPT-4o, Llama, and Mistral models for various tasks in the Sinhala language, revealing unsatisfactory results. Notably, offensive language detection was not attempted.

Prior work on offensive language detection has explored fine-tuning open-source LLMs like Llama and Mistral, primarily for high-resource languages like English (He et al., 2024; Christodoulou, 2024) and low-resource languages like Vietnamese (Truong et al., 2024). However, prior work has not explored open-source LLMs (e.g., Llama, Mistral) for Sinhala offensive language detection, despite their success in other low-resource languages like Vietnamese (Truong et al., 2024).

## 3 Method

### 3.1 Intermediate Pre-Finetuning Strategy

We adapt a two-stage fine-tuning strategy to optimize limited annotated data available for Sinhala. We train our models using the SOLD dataset (Ranasinghe et al., 2024) ($\mathcal{D}_{SOLD}$), which contains 7,500 training and 2,500 test samples. We split the training set into 9:1 (6,750 training, 750 validation) and reserve the test set for final evaluation. For more details on $\mathcal{D}_{SOLD}$, see Section 3.3.

Following Kim et al. (2022), we employ masked rationale prediction (MRP) as the intermediate task in the first stage of the fine-tuning strategy. For a sentence $S$, the embedded sentence can be represented as:

$$X^S = \left\{x_0^S, x_1^S, \ldots, x_{n-1}^S\right\} \in \mathbb{R}^{n \times d} \tag{1}$$

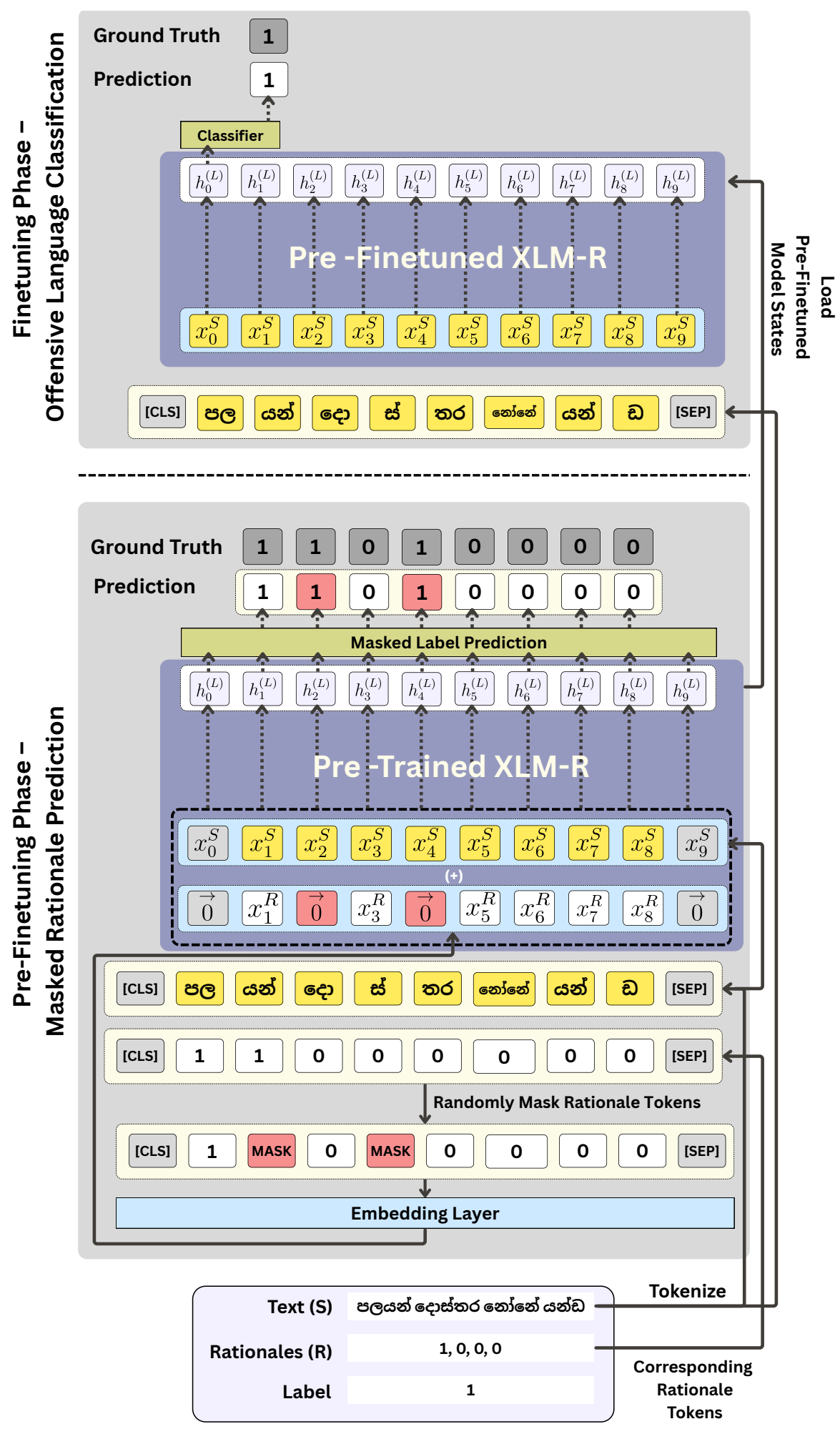

Figure 1: Two-stage fine-tuning strategy utilized to finetune a pre-trained `subasa-xlm-roberta-base` model.

where $n$ is the sequence length and $d$ is the embedding size. Similarly, the rationale labels $R$ can be represented as:

$$X^R = \left\{x_0^R, x_1^R, \ldots, x_{n-1}^R\right\} \in \mathbb{R}^{n \times d} \tag{2}$$

Unlike XLM-R's masked language modeling (MLM), which masks tokens, MRP masks rationale labels to construct partially masked rationale embeddings $\tilde{X}^R$. We randomly select and replace 75% of non-special rationale labels with zero vectors $\vec{0}$. For example, if $x_2^R$ and $x_4^R$ are masked:

$$\tilde{X}^R = \left\{\vec{0}, x_1^R, \vec{0}, x_3^R, \vec{0}, \ldots, x_{n-2}^R, \vec{0}\right\} \tag{3}$$

where the first and last tokens (CLS/SEP) are also zeroed. The model predicts masked rationale labels by combining $X^S$ with $\tilde{X}^R$:

| Hyper-parameter | Stage 1 | Stage 2 |
|---|---|---|
| Learning Rate | $2 \times 10^{-5}$ | $2 \times 10^{-5}$ |
| Batch Size | 16 | 16 |
| Epochs | 5 | 5 |
| Optimizer | RAdam | RAdam |
| Mask Ratio | 0.75 | - |
| Base Model | xlm-roberta-base | xlm-roberta-base |

Table 1: hyper-parameters for intermediate pre-finetuning and task-specific fine-tuning

$$H_{MRP}^{(0)} = X^S + \tilde{X}^R \tag{4}$$

$$H_{MRP}^{(l+1)} = \text{Transformer}\left(H_{MRP}^{(l)}\right) \tag{5}$$

$$\hat{X}^R = \text{MLP}\left(H_{MRP}^{(L)}\right) \tag{6}$$

Here, $H_{MRP}^{(l)}$ is the $l$-th transformer layer output, and $\hat{X}^R$ are predicted rationale labels.

**Stage 1 - MRP**: First we convert binary rationale labels (0/1 sequences) into padded tensors that align with the tokenized text length through rationale processing, ensuring dimensional compatibility with the input sequence.

These processed rationales undergo embedding fusion, where token embeddings $X^S$ (Equation 1) are combined with rationale embeddings $X^R$ (Equation 2) via summation to form the initialized representation $H_{MRP}^{(0)}$ (Equation 6). The fused embeddings then enter a masking phase, where 75% (selected as a hyperparameter for our implementation) of non-special tokens in $\tilde{X}^R$ (Equation 3) are randomly masked. We mask 75% of non-special tokens-a value empirically validated through ablation (Table 6) as optimal for balancing noise and learning signal for our Sinhala setting.

**Stage 2 - Offensive Language Detection**: Using the model states from Stage 1, we fine-tune for binary classification and train on the full $\mathcal{D}_{SOLD}$ training set. During both stages, we add special tokens (@USER, <URL>) to the tokenizer to handle frequent artifacts in training data.

Figure 1 provides an overview of the two-stage strategy described above, while Table 1 lists the hyperparameters used during both stages of the Intermediate Pre-Finetuning Strategy.

To contextualize our results, we compare against three baselines: (1) a **1D CNN** adapted from English sentiment analysis (Kim, 2014), (2) a **2D CNN** previously used for Sinhala NLP (Ranasinghe et al., 2019) (both using FastText (Bojanowski et al., 2017) embeddings), and (3) a vanilla fine-tuning of `xlm-roberta-base` on $\mathcal{D}_{SOLD}$. These represent traditional, domain-specific, and PLM-based approaches, respectively.

The performance of the models under the intermediate pre-finetuning strategy experiments is presented in Table 3.

#### 3.1.1 Ablation Study Design

To validate the impact of our intermediate Pre-Finetuning strategy, we conducted three ablation experiments using `xlm-roberta-base`:

1. **Masking Ratio Variation**: We trained models with MRP mask ratios $\in \{0.25, 0.75, 0.9, 1.0\}$, keeping all other hyper-parameters fixed (Table 1).

2. **Intermediate Task Replacement**: We replaced MRP with standard masked language modeling (MLM), using mask probabilities $\in \{0.15, 0.5\}$ and finetuned on $\mathcal{D}_{SOLD}$.

3. **No Intermediate Task**: Direct fine-tuning on $\mathcal{D}_{SOLD}$ without MRP/MLM, starting from the default `xlm-roberta-base` model states. Results are summarized in Table 6, with full metrics in Appendix Table 7.

### 3.2 Task Specific Fine-tuning Strategy

We instruction-finetune Llama-3 and Mistral models using parameter-efficient fine-tuning (PEFT) with 4-bit quantization (QLoRA). Our prompt (see Appendix A for the full prompt template) is structured for classification (`OFF`/`NOT`) and offensive phrase extraction, encouraging localization of offensive content. We employ LoRA (Hu et al., 2021) (rank=16, $\alpha$=16) targeting all linear projections, balancing efficiency and performance. Table 2 shows the list of hyper-parameters used during training for task specific fine-tuning.

**Training Data**: Using the prompt template (Appendix A) for each $\mathcal{D}_{SOLD}$ training sample, we populate the prompt with: The original Sinhala text in the '[TWEET]' field, The ground-truth label (`OFF`/`NOT`) in the '[LABEL]' field, and offensive phrases extracted from contiguous spans of rationale-annotated tokens in the '[PHRASES]' field. We validate the effectiveness of our fine-tuning strategy with the following baselines:

**Aya101** (Üstün et al., 2024) (multilingual instruction-finetuned) and **GPT-4o** are evaluated using the same prompt in zero-shot mode with the same prompt template. The performance of the models following task specific fine-tuning are presented in Table 4.

| **Hyper-parameter** | **Value** |
|---|---|
| Learning Rate | $2 \times 10^{-4}$ |
| Batch Size | 16 |
| Epochs | 5 |
| Optimizer | AdamW (8-bit) |
| Mask Ratio | 0.75 |
| Lora-R | 16 |
| Lora-Alpha | 16 |
| Lora-Dropout | 0 |
| Target Modules | `{ "q_proj", "k_proj", "v_proj", "o_proj", "gate_proj", "up_proj", "down_proj" }` |
| Max Sequence Length | 2048 |
| Per Device Train Batch Size | 4 |
| Gradient Accumulation Steps | 4 |
| Weight Decay | 0.01 |

Table 2: hyper-parameters for task specific fine-tuning

### 3.3 Dataset

We utilize $\mathcal{D}_{SOLD}$ (Ranasinghe et al., 2024), the largest publicly available dataset for identifying offensive language in the Sinhala script. Among the limited number of Sinhala offensive language datasets, $\mathcal{D}_{SOLD}$ stands out as the only one providing rationale labels, where `1` indicates a token that serves as a rationale for the offensive label, and `0` denotes a non-rationale token. A rationale can be defined as a specific text segment that justifies the human annotators decision of the sentence-level labels.

$\mathcal{D}_{SOLD}$ consists of data collected from Twitter and only contains tweets written in the Sinhala script, excluding those in Roman script or mixed script. Sentence-level offensive labels were determined by majority voting among the three annotators. Offensive tokens were identified based on agreement between at least two out of the three annotators, establishing the ground truth for token-level annotations (Ranasinghe et al., 2024). Selected examples from $\mathcal{D}_{SOLD}$ are given in Appendix Table 8.

From the original dataset, a random split was performed, where 75% of the instances were assigned to the training set, and the remaining instances were assigned to the testing set. We split the training set again into 9:1 (6,750 training, 750 validation) and reserve the testing set for final evaluation. Appendix figure 2 describes the class distribution in the dataset.

| Model | OFFENSIVE | | | NOT OFFENSIVE | | | Weighted | | | Macro |
|---|---|---|---|---|---|---|---|---|---|---|
| | P | R | F1 | P | R | F1 | P | R | F1 | F1 |
| 1D CNN Model (Kim, 2014) | 0.60 | 0.81 | 0.69 | 0.83 | 0.64 | 0.71 | 0.84 | 0.70 | 0.70 | 0.69 |
| 2D CNN Model based on Ranasinghe et al. (2019) | 0.79 | 0.65 | 0.69 | 0.79 | 0.85 | 0.82 | 0.78 | 0.78 | 0.77 | 0.76 |
| xlm-roberta-base-no-finetuning | 0.00 | 0.00 | 0.00 | 0.59 | 1.00 | 0.74 | 0.35 | 0.59 | 0.44 | 0.37 |
| xlm-roberta-base-vanilla-finetuned | 0.77 | 0.82 | 0.79 | 0.87 | 0.83 | 0.85 | 0.83 | 0.82 | 0.82 | 0.82 |
| Subasa-XLM-R | 0.78 | 0.84 | 0.81 | 0.89 | 0.84 | 0.86 | 0.84 | 0.84 | 0.84 | **0.84** |

Table 3: Evaluation results of **Subasa-XLM-R** and other baselines on $\mathcal{D}_{SOLD}$. We report per class (OFFENSIVE, NOT OFFENSIVE) precision (P), recall (R), and F1, and their weighted averages. Macro-F1 is listed with the best result in bold.

| Model | OFFENSIVE | | | NOT OFFENSIVE | | | Weighted | | | Macro |
|---|---|---|---|---|---|---|---|---|---|---|
| | P | R | F1 | P | R | F1 | P | R | F1 | F1 |
| Mistral-7b-instruct-v0.3 | 0.405 | 0.991 | 0.575 | 0.550 | 0.007 | 0.014 | 0.491 | 0.406 | 0.242 | 0.295 |
| Meta-Llama-3.1-8B-Instruct | 0.564 | 0.375 | 0.449 | 0.655 | 0.805 | 0.723 | 0.619 | 0.6315 | 0.612 | 0.586 |
| Meta-Llama-3.2-3B-Instruct | 1.000 | 0.000 | 0.000 | 0.594 | 1.000 | 0.745 | 0.758 | 0.594 | 0.443 | 0.373 |
| Aya101 (Üstün et al., 2024) | 0.864 | 0.422 | 0.567 | 0.707 | 0.954 | 0.812 | 0.771 | 0.738 | 0.713 | 0.690 |
| GPT-4o-2024-05-13 | 0.622 | 0.584 | 0.748 | 0.928 | 0.938 | 0.717 | 0.799 | 0.734 | 0.730 | 0.733 |
| Subasa-Mistral-7b-instruct-v0.3 | 0.917 | 0.611 | 0.734 | 0.783 | 0.962 | 0.863 | 0.838 | 0.820 | 0.811 | 0.799 |
| Subasa-Llama-3.2-3B | 0.822 | 0.698 | 0.755 | 0.813 | 0.896 | 0.853 | 0.816 | 0.816 | 0.813 | 0.804 |
| Subasa-Llama-3.1-8B | 0.837 | 0.738 | 0.785 | 0.834 | 0.902 | 0.867 | 0.836 | 0.836 | 0.834 | **0.826** |

Table 4: Evaluation results of **Subasa-Llama** and **Subasa-Mistral** and other baselines on $\mathcal{D}_{SOLD}$. We report per class (OFFENSIVE, NOT OFFENSIVE) precision (P), recall (R), and F1, and their weighted averages. Macro-F1 is listed with the best result in bold.

| Example | | GT | Our Models (Subasa) | | | Baselines | | | | | |
|---|---|---|---|---|---|---|---|---|---|---|---|
| Sinhala Text | Translation | | Llama3.1 | Mistral | XLM-R | GPT4o | Aya101 | Mistral | Llama3.1 | XLM-R-L | XLM-R-B |
| @USER පොහොට්ටුවේ උන්ගේ සැබෑ ස්වරූපය තමයි ඕක. අමු තිරිසන්නු | @USER That is the true nature of those in Pohottuwa. Real savages. | OFF | OFF | OFF | OFF | NOT | OFF | NOT | NOT | OFF | NOT |
| @USER ඒ දෙක පස්ස පැත්තෙ ගහගනිං | @USER stick those two up your ass. | OFF | OFF | NOT | NOT | NOT | NOT | NOT | NOT | NOT | NOT |
| "ඒ ගොනා වික්කා" කියලා කොහොමද ඉංග්‍රීසියෙන් කියන්නේ #askingforafriend | How do you say “I sold that bull” in English? | OFF | NOT | NOT | OFF | OFF | OFF | OFF | OFF | OFF | OFF |

Table 5: Classification examples from $\mathcal{D}_{SOLD}$ showing model predictions. Original Sinhala text with translations, ground truth (GT), our Subasa models' predictions, and baseline comparisons. **OFF**: Offensive, **NOT**: Non-offensive.

## 4 Results and Discussion

Concerning **RQ1**, our Subasa-XLM-R model achieves a macro-F1 of 0.84 (Table 3), outperforming both CNN baselines and the vanilla fine-tuned XLM-R (0.82 macro-F1). This 2% improvement demonstrates that MRP effectively bridges the gap between pre-training and downstream task adaptation in Sinhala's low-resource setting. The class imbalance in $\mathcal{D}_{SOLD}$ (Appendix 2) was the reason behind the use of macro-F1 for performance comparison, which equally weights both classes despite the majority NOT OFFENSIVE examples.

Ablation Study insights show that MLM with 50% masking matches MRP's performance (0.83 vs 0.84 macro-F1). This suggests that in low-resource settings, *any* token-level intermediate task (MLM/MRP) can enhance downstream performance by reinforcing local context understand-ing. While both MRP and MLM improve performance, their similar results warrant further study into task-specific intermediate objectives for low-resource languages.

Concerning **RQ2**, our results (Table 4) show significant gains across all LLM variants. The Subasa-Llama-3.1-8B model, derived from Meta-Llama-3.1-8B-Instruct, achieves the highest macro-F1 of 0.826, outperforming its base version (0.586 to 0.826). Similarly, Subasa-Llama-3.2-3B–adapted from Meta-Llama-3.2-3B-Instruct–achieves a macro-F1 of 0.804, more than doubling its base model's performance (0.373 to 0.804). The Subasa-Mistral-7B variant, built on Mistral-7B-Instruct-v0.3, also shows improvement compared to its base version (0.295 to 0.799). All our models surpass GPT-4o's zero-shot performance (0.733 macro-F1), with even the 3B Subasa-Llama model outperforming GPT-4o despite being a significantly smaller model. This highlights how task-specific fine-tuning with QLoRA enables open-source LLMs to specialize for low-resource languages.

When comparing results from Table 3 and Table 4, while Subasa-Llama-3.1-8B (0.826 macro-F1) leads among LLM variants, it slightly trails the smaller Subasa-XLM-R model (0.84 macro-F1). This counterintuitive result, where a 270M-parameter model outperforms an 8B-parameter LLM, suggests MRP's intermediate task provides a focused learning signal for offensive language detection, compensating for the XLM-R model's smaller size. Another factor is that the Subasa-Llama variants, despite their larger parameter count, inherit base models (Llama-3.1/3.2-Instruct) with minimal Sinhala pre-training data compared to XLM-R's multilingual foundation which contains the Sinhala language in its pre-training corpus.

## 5 Conclusion

This study addresses the challenge of offensive language detection in Sinhala, a low-resource language, by introducing four novel models: Subasa-XLM-R, Subasa-Llama (two variants), and Subasa-Mistral. To the best of our knowledge, our work is the first to adapt intermediate pre-finetuning and task-specific fine-tuning strategies for Sinhala, demonstrating significant advancements over existing baselines and state-of-the-art LLMs like GPT-4o. Below, we summarize our findings in relation to our initial research questions posed in Section 1:

***RQ1***: *Can intermediate pre-finetuning tasks (e.g., masked rationale prediction) improve PLMs for offensive language detection in Sinhala?* Our results confirm that intermediate pre-finetuning with MRP enhances model performance, with Subasa-XLM-R achieving a macro-F1 of 0.84, surpassing vanilla fine-tuned XLM-R (0.82). Ablation studies reveal that token-level intermediate tasks—whether MRP or standard MLM—improve downstream task performance for Sinhala (a low resource setting). Notably, MLM with 50% masking nearly matches MRPs gains (0.83 vs. 0.84 macro-F1), suggesting that reinforcing local context understanding through intermediate tasks aids the performance of the downstream task for Sinhala.

***RQ2***: *Can task-specific fine-tuning improve LLMs for offensive language detection in Sinhala?* Our results indicate that QLoRA enables open-source LLMs to specialize effectively for Sinhala and surpass GPT-4o's zero-shot performance. (e.g., Subasa-Llama-3.1-8B achieves a macro-F1 of 0.826, outperforming GPT-4o (0.733) and its base model (0.586).)

We publicly release all models and code to support Sinhala NLP research. Our results establish that strategic fine-tuning is beneficial for low-resource offensive language detection, with implications for other underrepresented languages.

| **Configuration** | **Accuracy** | **Macro F1** |
|---|---|---|
| **Intermediate Task = MRP** | | |
| Mask Ratio = 0.25 | 0.83 | 0.83 |
| Mask Ratio = 0.5 | 0.82 | 0.82 |
| Mask Ratio = 0.75 | **0.84** | **0.84** |
| Mask Ratio = 1.00 | 0.83 | 0.83 |
| **Intermediate Task = MLM** | | |
| Mask Prob = 0.15 | 0.84 | 0.83 |
| Mask Prob = 0.50 | 0.84 | 0.83 |
| **No Intermediate Task** | 0.82 | 0.82 |

Table 6: Ablation Study Results

## Limitations

In our approach, we adopted `xlm-roberta-base` as the foundation for Subasa-XLM-R due to hardware and computational resource limitations. This choice precludes direct comparisons with larger variants such as `xlm-roberta-large`, which might exhibit different behaviors when subjected to our intermediate pre-finetuning strategy. Similarly, our experiments with Mistral and Llama 3 models were restricted to smaller variants, limiting insights into how larger variants of these LLMs might perform in our task-specific finetuning strategy.

Our approach to the task-specific fine-tuning strategy utilized a single prompt template in a zero-shot prompting setting during training for consistency. While this approach reduced variability in experiments, it limited insights into the sensitivity of results against alternative prompting strategies.

## Acknowledgments

This work was supported by the Google Cloud Research Credits program (Award GCP19980904) and the OpenAI Researcher Access Program with API Credits. We would like to thank the anonymous reviewers for their valuable feedback and suggestions. Additionally, we thank Nate Raw for their insightful discussions.

## A Prompt Template

The full instruction template used for fine-tuning is shown below:

```
System: "You are an emotionally
   intelligent assistant who
   speaks Sinhala and English
   Languages. Your task is to
   determine whether each tweet
   is OFFENSIVE or NOT OFFENSIVE.
    For each tweet, provide a
   single word as your output:
   either \"OFF\" or \"NOT\". For
    offensive tweets, identify
   and list the specific
   offensive phrases without
   translation.\n"

User: "Please classify the
   following tweet as \"OFF\" or
   \"NOT\". If offensive, list
   the specific offensive phrases
   :\n\n'[TWEET]'"

Assistant: "[LABEL]\nPhrases: [
   PHRASES]"
```

**Placeholders**: - [TWEET]: Original Sinhala text from $\mathcal{D}_{SOLD}$. - [LABEL]: Ground-truth label (OFF or NOT). - [PHRASES]: Offensive phrases extracted from rationale annotations.

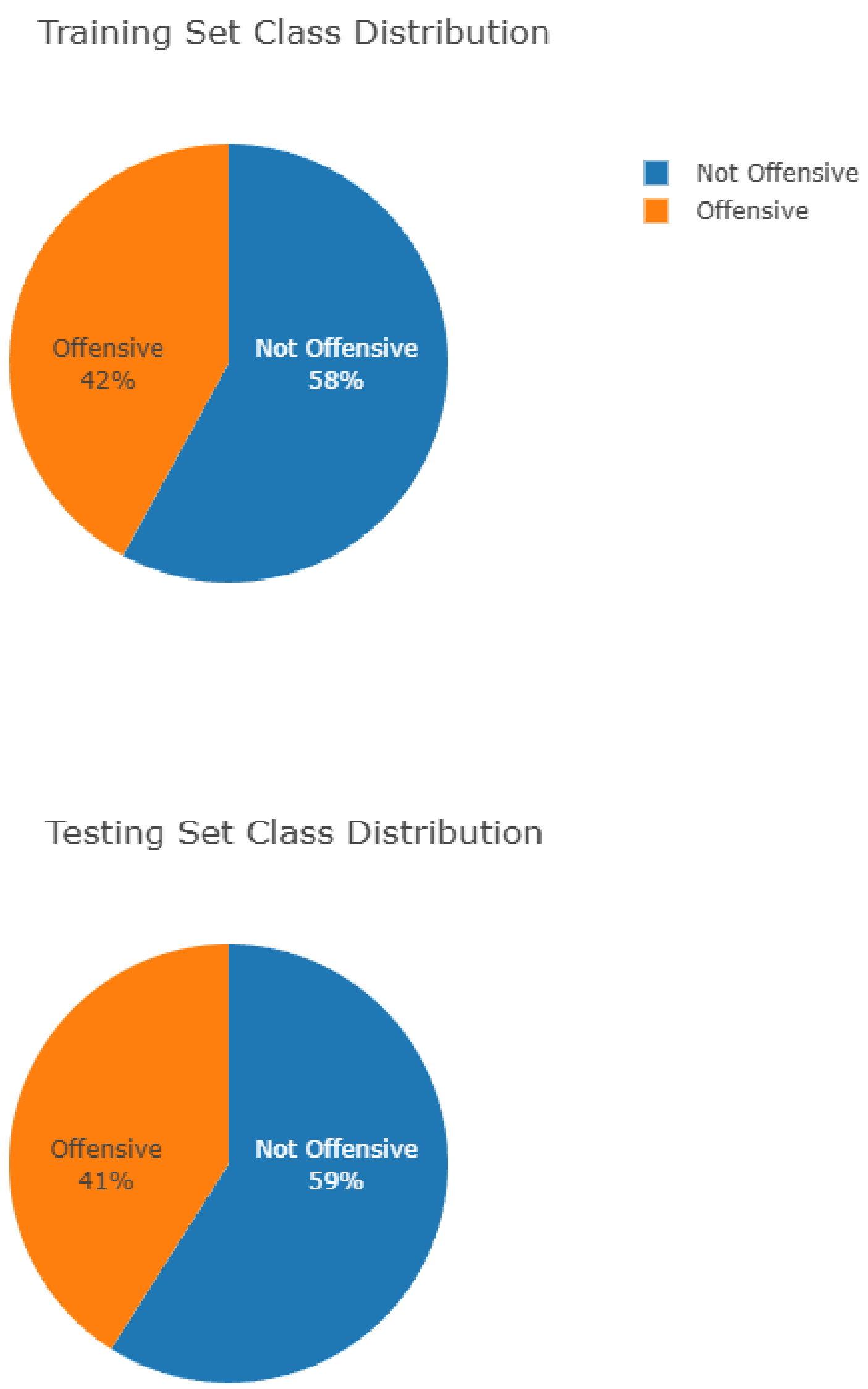

Figure 2: Class Distribution of Training and Testing Sets: The pie charts illustrate the distribution of 'NOT Offensive' and 'Offensive' instances in the training set (75% of the original dataset) and testing set (25% of the original dataset). $\mathcal{D}_{SOLD}$ contains 10,000 Sinhala tweets in total, and out of these 4191 are labeled as offensive and 5,809 labelled as non-offensive.

| Configuration | OFFENSIVE | | | NOT OFFENSIVE | | | Weighted | | | Macro |
|---|---|---|---|---|---|---|---|---|---|---|
| | P | R | F1 | P | R | F1 | P | R | F1 | F1 |
| **MRP (Ours)** | | | | | | | | | | |
| Mask Ratio = 0.25 | 0.79 | 0.82 | 0.80 | 0.87 | 0.85 | 0.86 | 0.84 | 0.84 | 0.84 | 0.83 |
| Mask Ratio = 0.50 | 0.85 | 0.72 | 0.78 | 0.83 | 0.91 | 0.87 | 0.83 | 0.83 | 0.83 | 0.82 |
| **Mask Ratio = 0.75** | 0.79 | 0.85 | 0.82 | 0.89 | 0.84 | 0.87 | 0.85 | 0.85 | 0.85 | **0.84** |
| Mask Ratio = 1.00 | 0.78 | 0.81 | 0.80 | 0.87 | 0.84 | 0.85 | 0.83 | 0.83 | 0.83 | 0.83 |
| **MLM Intermediate** | | | | | | | | | | |
| Mask Prob = 0.15 | 0.81 | 0.80 | 0.81 | 0.87 | 0.87 | 0.87 | 0.85 | 0.85 | 0.85 | 0.84 |
| Mask Prob = 0.50 | 0.82 | 0.79 | 0.80 | 0.86 | 0.88 | 0.87 | 0.84 | 0.84 | 0.84 | 0.84 |
| **No Intermediate Task** | 0.77 | 0.82 | 0.80 | 0.87 | 0.83 | 0.85 | 0.83 | 0.83 | 0.83 | 0.82 |

Table 7: Complete ablation study results on `XLM-R-Base` with per-class metrics. All experiments used identical training data, validation splits, and hyperparameters (Table 1). We report Precision (P), Recall (R), and F1 for both classes, along with weighted averages and Macro-F1. Best MRP configuration (Mask Ratio = 0.75) shown in bold.

| Tweet | Human Translation | Label | Rationales |
|---|---|---|---|
| @USER ඒ හිතන් ඉන්නේ @USER වගේම මටත් මෝඩ විමසම් කියලා .සැමක් mate. | @USER She thinks that I get aroused like her. Poor thing mate. | NOT | [] |
| @USER @USER නේ. ඇය ඉස්සර විචාරක කෙනෙක්? | @USER @USER Damn, isnt this the girl who used to be a news anchor | NOT | [] |
| @USER .. . එන්න ඔබේ ජන්ම දා ඌ * පරීක්ෂා කරනවා | @USER @USER Yo.. do you like to get your a** cracked open on your birthday | OFF | [0, 0, 0, 0, 0, 0, 0, 0, 1, 0, 1, 0, 0, 0] |

Table 8: Examples from $\mathcal{D}_{SOLD}$.